\documentclass{article}

\usepackage{PRIMEarxiv}

\usepackage[utf8]{inputenc} % allow utf-8 input
\usepackage[T1]{fontenc}    % use 8-bit T1 fonts
\usepackage{hyperref}       % hyperlinks
\usepackage{url}            % simple URL typesetting
\usepackage{booktabs}       % professional-quality tables
\usepackage{amsfonts}       % blackboard math symbols
\usepackage{nicefrac}       % compact symbols for 1/2, etc.
\usepackage{microtype}      % microtypography
\usepackage{lipsum}
\usepackage{xcolor}
\usepackage{soul}
\usepackage{amsmath}
\usepackage{fancyhdr}       % header
\usepackage{graphicx}       % graphics
\usepackage{ifluatex}
\usepackage{setspace}
\usepackage{algorithm}
\usepackage{algpseudocode}
\usepackage{mathtools}
\usepackage{tablefootnote}
\usepackage{amssymb}
\usepackage{subcaption}

\ifluatex
  \usepackage{pdftexcmds}
  \makeatletter
  \let\pdfstrcmp\pdf@strcmp
  \let\pdffilemoddate\pdf@filemoddate
  \makeatother
\fi

\usepackage[notransparent]{svg}

\renewcommand\hl[1]{#1}

\graphicspath{{media/}}     % organize your images and other figures under media/ folder

\title{Evaluation and Comparison of Federated Learning Algorithms for Human Activity Recognition on Smartphones \thanks{This work has been partially funded by Naval Group, and by MIAI@Grenoble Alpes (ANR-19-P3IA-0003). This work was also granted access to the HPC resources of IDRIS under the allocation 2022-AD011013233 made by GENCI.}

}

\author{Sannara EK \\ sannara.ek@univ-grenoble-alpes.fr \\ University Grenoble Alpes \\ LIG F-38000 \\ Grenoble, France 
\And
\textbf{François PORTET} \\ francois.portet@univ-grenoble-alpes.fr  \\ University Grenoble Alpes \\ LIG F-38000 \\ Grenoble, France 
\And
\textbf{Philippe LALANDA} \\ philippe.lalanda@univ-grenoble-alpes.fr \\ University Grenoble Alpes \\ LIG F-38000 \\ Grenoble, France 
\And
\textbf{German VEGA} \\german.vega@univ-grenoble-alpes.fr \\ University Grenoble Alpes \\ LIG F-38000 \\ Grenoble, France}

\begin{document}
\maketitle

% \tnotetext[t1]{This work has been partially funded by Naval Group, and by MIAI@Grenoble Alpes (ANR-19-P3IA-0003).}

\begin{abstract}
%% Text of abstract
Pervasive computing promotes the integration of smart devices in our living spaces to develop services providing assistance to people. Such smart devices are increasingly relying on cloud-based Machine Learning, which raises questions in terms of security (data privacy), reliance (latency), and communication costs. In this context, Federated Learning (FL) has been introduced as a new machine learning paradigm enhancing the use of local devices. At the server level, FL aggregates models learned locally on distributed clients to obtain a more general model. 
In this way, no private data is sent over the network, and the communication cost is reduced. Unfortunately, however, the most popular federated learning algorithms have been shown not to be adapted to some highly heterogeneous pervasive computing environments. In this paper, we propose a new FL algorithm, termed FedDist, which can modify models (here, deep neural network) during training by identifying dissimilarities between neurons among the clients. This permits to account for clients' specificity without impairing generalization.
FedDist evaluated with three state-of-the-art federated learning algorithms on three large heterogeneous mobile Human Activity Recognition datasets. Results have shown the ability of FedDist to adapt to heterogeneous data and the capability of FL to deal with asynchronous situations.

\end{abstract}

% %%Graphical abstract
% \begin{graphicalabstract}
% %\includegraphics{grabs}
% \end{graphicalabstract}

%%Research highlights
%\begin{highlights}
%\item Research highlight 1
%\item Research highlight 2
%\end{highlights}

%% \linenumbers

%% main text
\section{Introduction}

Pervasive computing relies on the integration of smart devices in our living spaces to support the development of services to individuals and organizations \cite{7488250,pervasiveTrend}. For several years, we have seen the emergence of smarter services based on Machine Learning (ML) techniques. ML enables the production of highly performing decision systems by identifying patterns that may be hidden within massive data sets whose exact nature is unknown and therefore cannot be programmed explicitly. Bringing such services into production nevertheless raises major architectural problems. According to most current solutions, ML models are built in the cloud using historical data, then deployed and executed on devices or on the edge. Additional data is then regularly collected by the devices and sent to the cloud in order to build new and more up-to-date models. Such an approach, however, undergoes major limitations in terms of security, performance, resilience and even cost. 
% \cite{diao2021heterofl}

\begin{figure}[!htb]
\centering
\includegraphics[width=0.75\linewidth]{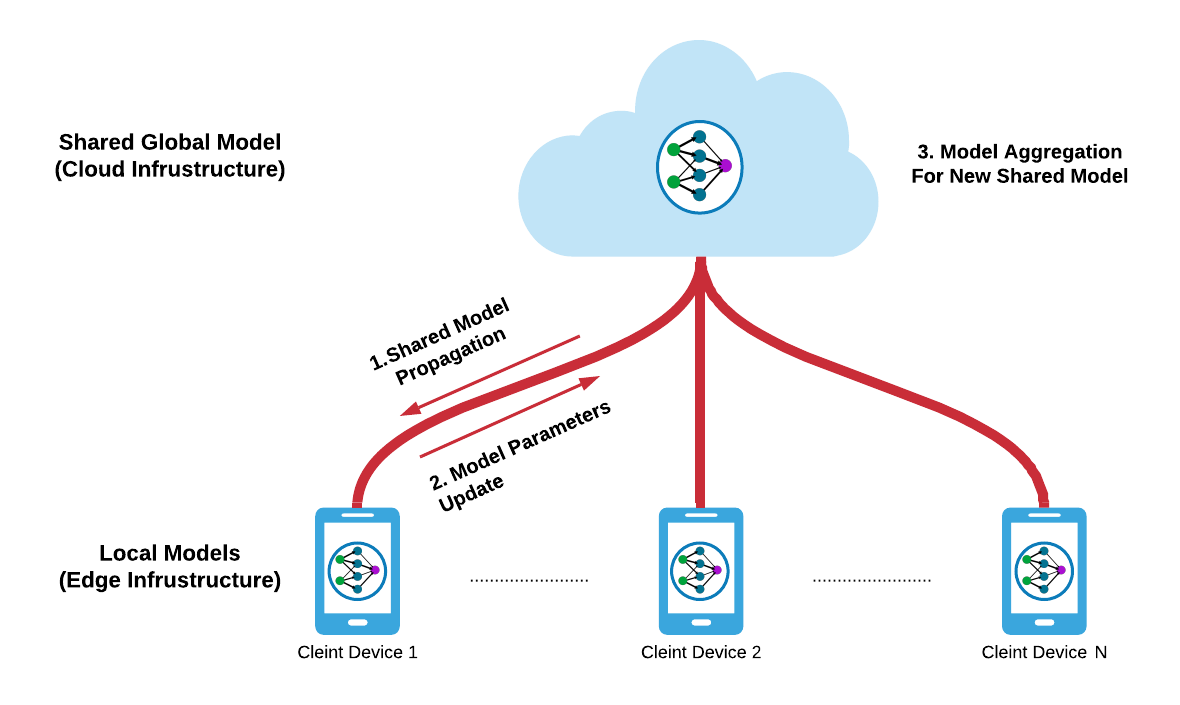}
\caption{Federated learning architecture and principles.}
\label{fig:FL_principles}
\end{figure}

Federated learning, depicted in figure \ref{fig:FL_principles}, was recently proposed  \cite{mcmahan2016communicationefficient,DBLP:journals/corr/abs-1902-01046,DBLP:journals/corr/KonecnyMRR16} as a new machine learning paradigm promoting the execution and specialization of local models on devices and their regular sending to a server where they are aggregated into a more generic one. The new model is redistributed to devices for the next local learning iteration. Federated learning reduces communication costs and improves security because no private data (only models) is exchanged \cite{Lim2020}. It has immediately attracted attention in the pervasive field but, so far, it has been essentially used in traditional learning fields like computer vision \cite{mcmahan2016communicationefficient,fedperr,li2019fedmd,wang2020federated} and still needs to be adapted to the specificity of the pervasive domain. Originally, FL was set out as an alternative form of distributed learning to obtain a single well-performing model on the server-side. However, to deal with pervasive computing, this learning approach should shift from a single server-centric model to a multi client-centric model objective \cite{9415623,lee2021opportunistic}. The client models should perform exceedingly well, compared to a single conventionally trained generalized model, on their local data (\emph{strong personalization}) and be able to deal with unseen data compared to approaches without any collaborative training (\emph{good generalization}). To reach this goal, challenges are numerous.
In particular, the high \emph{heterogeneity} of the devices is problematic. For instance, smartphones have different brands, sensors, ways of being used which makes data collected in this context challenging for machine learning. It is also well known that mobile  devices evolve at different paces and can have very unreliable network connections, which involves synchronization issues. The effects of \emph{asynchronicity} on FL need investigations \cite{Lim2020}.

In this paper, we introduce a new aggregation algorithm, called FedDist with the implementation available on github \footnote{PerCom2021-FL - \url{https://github.com/getalp/PerCom2021-FL}}, that has been designed to meet the specific needs of pervasive applications. This algorithm, together with state-of-the-art FL methods, has been assessed in an extensive set of experiments, based on a well-defined evaluation method. These experiments are all performed in the illustrative field of Human Activity Recognition (HAR) on smartphones. This domain aims to automatically identify human physical activities, like running or walking, using sensors embedded in smartphones. HAR is well suited to our FL experiments because activities tend to have generic patterns while being highly idiosyncratic (data depends on people, devices, the way devices are carried, the environment, etc.) \cite{10.1145/2809695.2809718,ek2020evaluation}. These experiments have been performed with large, realistic and freely available data to study both the classification performance of the FL algorithms with heterogeneous data and their ability to deal with devices that come and go during the learning process. The results exhibits the relevance of the FL approach for the latter and the superiority of FedDist for the former. Part of this work has been published in \cite{ek:hal-03207411}, but this paper presents an updated version with more experiments and a new study on the effect of asynchronicity.

%evaluated is  We extensively organize sets of experiments that each offer different insights on different algorithms in contrasting environments that use a straightforward evaluation method to measure and compare FedDist and three other representative algorithms against conventional approaches. Furthermore, asynchronous client simulations are introduced to replicate real-world scenarios where user participation within a federated learning communication round is not predictable and non-deterministic.}

%\hl{I changed the below to the above}

%\hl{
%In this paper, we propose a new aggregation algorithm, called FedDist, that has been designed to meet the specific needs of pervasive applications. We also present an extensive set of experiments, based on a well-defined evaluation method, that has been conducted to assess FedDist, as well as three other representative algorithms.}

%Experiments were carried out in the 
%Furthermore, the collected data is private and should not be sent over the network. HAR has long been addressed as a classification problem where the common approach is to process windows of data streams to extract a vector of features that, in turn, is used to feed a classifier. Many instance-based classifiers such as Bayesian Network, Random Forest, or Support Vector Machines have been used with reasonable success \cite{6181018,blachon2014}. Today, however, the most popular and effective technology is undoubtedly deep neural networks \cite{IGNATOV2018915,Cho}.

The paper is organized as follows. First, some background about FL and its most representative algorithms is provided. Then, our proposed algorithm, FedDist, is detailed in Section~\ref{sec:feddist}. Section~\ref{sec:eval} introduces the method that has been defined to evaluate and compare the different aggregation algorithms. Section~\ref{sec:expe} presents experiments performed with three public HAR datasets. This section also reports a study on the effect of asynchronicity on FL performances. Finally, the paper ends with a conclusion based on our main findings and an outlook on future work.

\section{Background}\label{sec:sota}

\subsection{Federated Learning principles}

A core incentive of FL in pervasive computing is to utilize mutual learning goals between clients through collaborative learning to share beneficial knowledge with one another. However, despite clients sharing the same goal, effective collaboration in the wild between unmoderated clients, as exemplified in cross-device and vertical FL settings \cite{li2021survey}, is an open challenge for any collaborative learning framework. The cause is widely known to be due to statistical heterogeneity (difference between client data caused by differences in individual user's usage) and system heterogeneity (difference between client data caused by different system traits) \cite{10.1145/2809695.2809718}. %There have been numerous studies that have proposed intricate methods to mitigate the detrimental effects of heterogeneity.

{
There have been many implementations of FL proposed by the scientific community to tackle heterogeneity. For instance, FedProx {\cite{li2020federated}} and FSVRG {\cite{konecny2016federated}}  apply local model/gradient controlled learning that limits the divergence of clients. Other approaches, such as FeSEM {\cite{xie2020multicenter}} and MLMG {\cite{9431011}} cluster clients or create more than one server model based on the clients' similarity. Finally, some FL aggregation algorithms consider local models in their entirety (all layers and neurons) and builds a new model that potentially calls into question all layers and all weights associated with neurons. This approach is shown in FedAvg {\cite{mcmahan2016communicationefficient}}  and FedMA {\cite{wang2020federated}} algorithms. Here our study focuses on this last family of solutions to deal with heterogeneity that emphasizes aggregation techniques at the server level and tries to address the  question: ``How can we effectively combine client models at the server level to create a global model that fits all the clients?''}

{In this study, we compare FedDist to three representative algorithms, namely FedAvg, FedProx and FedMA.}

   \textbf{FedAvg \cite{mcmahan2016communicationefficient}} is the pioneering FL algorithm. The aggregation is performed in a weighted averaging manner that make clients with more data having more weight in the newly aggregated model. FedAvg, however, has a naive form of aggregation due to its coordinate-wise averaging that may lead to sub-optimal solutions. For instance, with non-Independent and Identically Distributed (non-IID) data, neurons in the same coordinate may be opted for different purposes due to clients' specialization \cite{wang2020federated}. 
   
   \textbf{FedProx {\cite{li2020federated}}} adds a proximity term to the learning function of clients to limit the divergence of clients from the server model. This regularization technique allows the local learning to be done with respect to the server model, which allows a more harmonized aggregation for the next server model. FedProx demands a trade-off between personalization and generalization improvement in client model performances that can slow down the global convergence rate.
   
   \textbf{FedMA \cite{wang2020federated}} is an FL algorithm that modifies the neural model architecture by incorporating a layer-wise aggregation process where similar neurons can be aggregated and dissimilar ones can be added as new neurons in the architecture. This approach treats the number of nodes in a layer as a sub-problem to solve rather than a hyper-parameter to be set \cite{pmlr-v97-yurochkin19a}. FedMA considers that neurons in a neural network layer are permutation invariant to perform a more intelligent aggregation. The neurons (or filters) can be clustered in a non-parametric way where all neurons in the same cluster are averaged to produce a global neuron. In the search for which neurons can be fused, the algorithm uses a 2D permutation matrix that is computed iteratively from increasing layers level. Our experiments, however, found that calculating the permutation matrix makes the algorithm particularly slow during the aggregation stage.

\subsection{Synthesis}

Federated learning is a very promising approach but still in its infancy. A major issue today is the lack of structured, extensive tests of the different aggregation algorithms that have been proposed. So far, these algorithms are evaluated independently on traditional learning fields like computer vision, usually on relatively homogeneous data, with no outliers or little divergence at the client-side, although FL has been designed to tackle unevenly distributed data\cite{konecny2016federated}. %Also, performances are reported on known data and not on data that has not been seen during training. 
Also, most studies do not include the influence of the network or consider the client's behavior. 
Further studies are needed to really understand how the server and client models evolve and how well they behave regarding generalization (can a client model work correctly on data that have not been seen beforehand?) and specialization (can a client work well with data that have the same properties as the one used for training?). 

Federated learning has not been widely used in the context of pervasive computing. For instance, there are only a few works in the HAR domain \cite{wu2020personalized,8672262,8885054,NIPS2017_7029,9076082,Berlo2020}, with missing analysis regarding the performance of global and local models on generalization and specialization with different approaches. Additionally, most tests were performed on small and pre-processed datasets collected in laboratory environments. {Also, existing algorithms do not consider the specific requirements of pervasive computing, where data is usually heterogeneous and the learning problem may change over communication rounds as clients drop in and out of the learning pool.} For instance, FedAvg is bounded to a single model shape with a fixed amount of filters or neurons to tackle an ever-growing problem (since new data and clients keep coming). Finally, contrary to centralized learning, where the model is learned to increase performance on one unique dataset, in our vision of FL the objective is to increase all individual clients' performances. To do so, FL must account for the fact that data, even though in the same domain, might evolve very differently according to the clients and would necessitate adapting the features to be induced (i.e., the representation must be learned again).  In order to account for the heterogeneity of clients, a solution must be found to adapt the global model in a way that makes it respect the peculiarity of each client. The FedDist algorithm presented below is our attempt to reach this goal.

\section{FedDist, A Dynamic Model Growth Algorithm In Federated Learning}\label{sec:feddist}

In this section, we present a novel neuron matching algorithm for FL based on a euclidean distance dissimilarity measurement. This algorithm, which includes some elements of FedAvg and FedMA, is called FedDist (Federated Distance) for its emphasis on computing distances of neurons of similar coordinates when comparing clients and server models. 

% on data or class distribution skew between clients are handled by leveraging the native properties of FedAvg where aggregation is done in a weighted averaging manner (Clients with more data contribute more to the global model).}
% \francois{I don't understand this sentence}

FedDist recognizes, like FedMA, that some client's models may diverge because of heterogeneous data. This results in neurons that cannot be matched with neurons from other models (because of weights that are too far apart). A naive coordinate averaging approach, like in FedAvg, has very negative effects. Indeed, these diverging neurons are simply erased by the averaging process, while they are actually able to deal with specific situations not encountered by other clients. FedDist also recognizes that the model's structure is relatively stable, which means that neurons with the exact coordinates play a similar role. This view provides an opportunity to build on a coordinate-wise approach.

% The goal of FedDist does not concentrate on data or class distribution skew between clients. It is although handled by leveraging the native properties of FedAvg (Which FedDist extends upon) where aggregation is done in a weighted averaging manner. (Clients with more data contributes more to the global model)

\begin{figure}[!htb]
\centering
\includegraphics[width=0.75\linewidth]{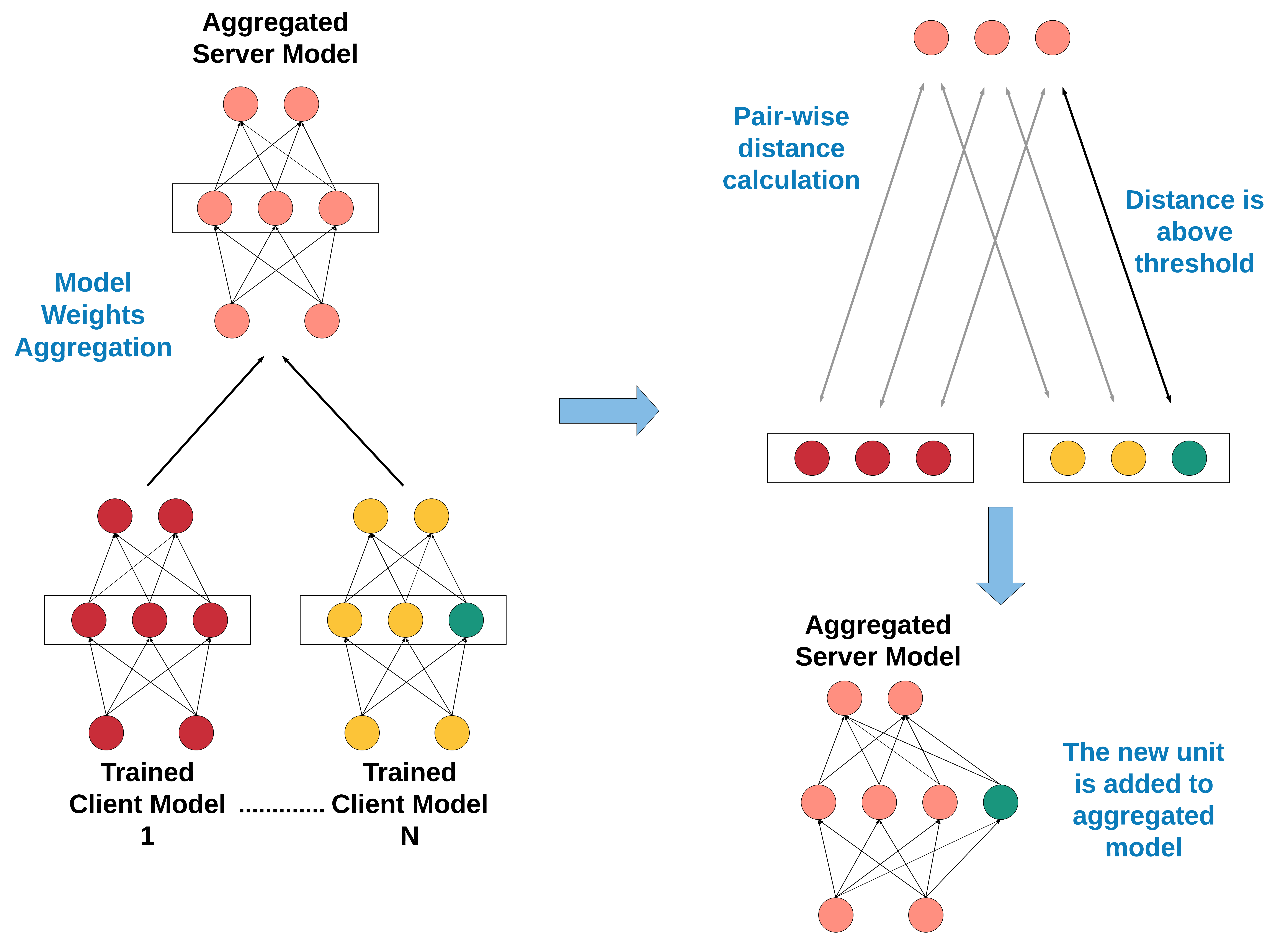}
\caption{FedDist unit generating process}
\label{fig:fedDistProcess}
\end{figure}

FedDist identifies diverging neurons using euclidean distances. These neurons that are specific to certain clients are added to the aggregated model as new neurons.  We believe that this technique is particularly suited for tackling sparse data, where specific features are only found in a small subset of clients or data points. Thus, this new neuron adding scheme can lead to larger models that are able to generalize better. As new neurons are added to a layer, a layer-wise training round is added in order to allow the neurons in the next layers to adjust to the new incoming neurons and weights. To do so, the layer with the new neuron and those below are frozen, and the subsequent layers are trained. This introduces intermediate communication rounds: a full communication round is finished when all layers have been treated, and a new aggregated model is computed. This process is summarized in Figure~\ref{fig:fedDistProcess}. First, a global server model is computed from all clients (left part of the figure). Outliers are identified using the euclidean distance (upper-right) and added to the aggregated model (lower-right) sent back to the client. 

The detailed algorithm is presented hereafter (Algorithm~\ref{fedDistAlgo}). It starts by distributing the server model ($w_t$) to all clients ($w_{tk}$). These clients then commit to train their model locally, where we denote the process as  $ClientUpdate\_start$ function, and send it to the server where a weighted averaging is performed similarly to FedAvg.  A pairwise euclidean distance (see equation~\ref{eq:l2} in section \ref{sec:eval} for details) is iteratively calculated for each unit in a layer between the client models and the aggregated server model to generate a cost-distance matrix ($\prod$).

% \begin{spacing}{0.8}
% \begin{algorithmic}
% \IF{$n$ is odd}
% \RETURN \TRUE
% \ELSE
% \RETURN \FALSE
% \ENDIF
% \end{algorithmic}
% \end{spacing}

\begin{algorithm}
\caption{Federated Distance (FedDist)}

\label{fedDistAlgo}
\begin{spacing}{1.0}
\begin{algorithmic}[1] 
\Require L = ModelLayerCount, T = CommunicationRound, K = ClientCount
\State initialize $\emph{w\textsubscript{1} on server}$
\For{each communication round $t = 1, 2, \cdots T$}
% \For{$t = 1, 2, \cdots T$}
    \For{$k = 1, 2, \cdots K$}
        \State $w_{tk} = w_{t}$
        % \State $ w_{nk}^l \Leftarrow localTrain(w_{nk}^l)$
        \State $w_{tk} \gets$ k.ClientUpdate\_start($w_{tk}$)
        % \Comment{Freeze layers below $l$}
    \EndFor
    \State $w_t \gets$  $\sum_{k=1}^{K}$  $\dfrac{n_k}{n}$  $w_{tk}$
    \vskip 3pt
    \For{each layer $l = 1, 2, \cdots L-1$}
    % \For{$l = 1, 2, \cdots L$}
        \For{each client $k = 1, 2, \cdots K$}
        % \For{$k = 1, 2, \cdots K$}
            \State $\prod_{t}^l \Leftarrow calculatePairWiseDistance(w_{t}^l,w_{tk}^l)$
        \EndFor
        \State $\mu^l,\sigma^l \Leftarrow calculateMean\&stdOfNeuron(\prod_{t}^l)$
        \State $newNeuron = False$
        \For{each neuron distance $d = 1, 2, \cdots D$ in $\prod_{t}^l$}
        \vskip 2pt
            \State $threshold = (penaltyFunc(t) + 3)*\sigma_{d}^l + \mu_d^l $
            \If{$mean(d) > threshold $}
                \State $appendNeuronTo(w_{t}^l)$
                \State $newNeuron = True$
            \EndIf
        \EndFor
        \If{$newNeuron$}
            \For{each client $k = 1, 2, \cdots K$}
                \State $w_{tk}^l = w_{t}^l$
                                        \Comment{Freeze layers $l$ and below}
                \vskip 2pt
                \State $w_{tk}^l \gets$ k.ClientUpdate\_start($w_{tk}^{l+1} ,...., w_{tk}^{L}$)
                \vskip 2pt
                \State $w_{t} \gets$  $\sum_{k=1}^{K}$  $\dfrac{n_k}{n}$  $w_{tk}^l$
            \EndFor
        \EndIf
    \EndFor
    \State $w_{t+1} \gets$  $w_{t}$
\EndFor

\end{algorithmic}
\end{spacing}
\end{algorithm}

To evaluate how different neurons are between the server and the client models, we proposed a \emph{Pair-wise Dissimilarity measurement}. When the client models are very close, the distance between neurons of different clients should be relatively low. In this work, we used an equivalent of the L2 norm on the difference between two vectors, that is the Euclidean distance between the weight of two neurons as below:

\smallskip
\begin{equation}\label{eq:l2}
\scriptstyle {dist(N_1,N_2) = \sqrt{ (N_1^{w_1} - N_2^{w_1})^2 + \dots + (N_1^{w_K} - N_2^{w_K})^2 }}
\end{equation}
\smallskip
where $N_1$ and $N_2$ are two different neurons and $w_i$ is the $i^{th}$ of the $K$ weights of the neurons. A large distance would indicate a strong dissimilarity, while neurons that are very similar to one another should have a small distance. {The method works similarly for convolutional filters, the only difference is the expansion of the weights by one dimension for the filter maps.}

% \begin{equation}\label{eq:l2}
% \scriptstyle {dist = \sqrt{ (N_1^{w11} - N_2^{w11})^2 + (N_1^{w21} - N_2^{w21})^2 + .... + (N_1^{wnk} - N_2^{wnk})^2 }}
% \end{equation}
% \smallskip
% where $N_1$ and $N_2$ are two different neurons and $wnk$ is the $n^{th}$ weight of the $k^{th}$ neuron. A large distance would indicate a strong dissimilarity, while neurons that are very similar to one another should have a small distance.

%Such measure is particularly useful to assess how diverging the client weights are from the server ones. This enables to identify clients that become outliers during the learning by constantly diverging from the server model. This measure is also the basis of the distant measure used in FedDist. 

%In this work, we have calculated the pair-wise distance between each client model and the server model for each communication round. In the coming section, results are presented in two different ways. First, the averages of the distances by layers are presented to exhibit where dissimilarity is highest in the architecture of the network. Secondly, the average for each individual client with the server model is presented to identify clients that become outliers.

Then, the mean ($\mu$) and standard deviation ($\sigma$) of the euclidean distance for every neuron are calculated. The average provides information about the direction taken by most clients. With the normal distribution property\footnote{{The normal distribution has been experimentally assessed from the statistics of the distance values during the communication rounds of a specific experiment.}}, the values less than three standard deviations away from the mean holds for 99.73\% of the set. Hence, we define a threshold $threshold = (penaltyFunc(currentRound,\beta)+ 3) * \sigma + \mu + $, where every value above three standard deviations are considered sufficiently specialized from the other to be added in the network. 
%
%With the normal distribution property, the values less than one standard deviation away from the mean holds for 68.27\% of the set, while two standard deviations holds for 95.45\% and three standard deviations for 99.73\%. Thus, by using this distribution property, we define a threshold as below:
%\begin{equation}\label{eq:treshold}
%threshold = 3*\sigma + \mu + penaltyFunc(comRound,beta)
%\end{equation}
% $threshold = 3*\sigma + \mu + penaltyFunc(comRound,beta)$
%
In order to limit the number of neurons added, the penalty function $penaltyFunc$ takes parameter $\beta$ that sets the increase rate of the threshold as training continues. %If a neuron in any of the clients holds an individual distance above the threshold, it is then added to the server model.  
The process is performed layer-wise. At each communication round, it is performed on the first layer. Then the first layer is updated and frozen at the client-side. The model is then retrained on the client-side (all already treated layers frozen) to allow the next layers to adjust and adapt to any newly added neuron weights. In this stage, the client model only needs to send back unfrozen layers of the model to reduce communication overhead. 

% REMOVED com overhead
If we consider all layers of equal size, the average communication overhead with respect to FedAvg is thus $\frac{(L-1)}{2}*FedAvg_{cost}$ since for $L$ layers, the process is repeated $L$ times with a decreasing number of layers. In practice, the cost is much lower since layer-wise training is skipped for layers where no unit has been added in the previous layer. Additionally, if all client models become fully saturated with no new neuron added, then layer-wise training is no longer needed, and the learning becomes equivalent to FedAvg.

% The server then sends the newly modified model to the clients for retraining with the layer processed and below frozen. 

% This is crucial as the subsequent neurons of the next layer requires fine tuning to adjust to any new weight from the neuron we added on the current layer. This step is an intermediate communication round, where a full communication round in FedDist concludes when clients finish training the last layer with all weights below it frozen and all models has been aggregated to the server.

%We know that clients are individually converging to their local-minima due to personalization during the on-device training stage. This may result in clients differing widely in the direction of the models. This can be better observed through the difference of filters/units which were of the same origin between the clients. In specific cases, the neurons in some clients may be drastically different that decremental effects would be obtained if used in the averaging with the pair neuron in other clients.

%A possible approach to this problem is to remove these outlier-like neuron from the aggregation with its counterpart in other clients. However, as this outlier like neuron may be a less popular feature not found on the majority of clients, it may be beneficial to keep the neuron to further the ability of our model to generalize by having it added to the global shape on the server model. 

\section{Evaluation Method}\label{sec:eval}

%A proper reproducible evaluation of FL algorithms requires a clear definition of the evaluation method and baseline systems. 
% This section describes the datasets considered in the study, the evaluation strategy, and the evaluation metrics. 

\subsection{HAR Task, Datasets Selection and Preparation} \label{sec:datasets}
In this paper, we favor reproducibility, heterogeneity, and realistic situations. Hence, the ideal dataset(s) should be: freely accessible, acquired in real-life environments with several participants and devices, include high-class imbalance, and carefully annotated. Furthermore, since FL implies several local learning phases, the dataset should be large enough to simulate an extended period of time. 

Despite the HAR task being well investigated, attempts to benchmark it on smartphones have only recently emerged. As reported in a recent survey \cite{Sussex_challenge}, a large number of datasets acquired from smartphones, worn in different ways, with various sensors and sampling frequency, make it difficult to reach a uniformity in tasks, sensors, protocols, time windows, etc. Furthermore, some datasets are very imbalanced because activity distributions among classes are very different. HAR is thus a perfect fit for testing FL in realistic scenarios. 

In the mobile HAR domain, a well-known dataset is the UCI dataset \cite{Anguita2013APD} which has been widely used as a benchmark in the domain. However, this dataset is not realistic (it was acquired in-lab following strict scenarios), and it is small in size (3.6 hours). Since the size and diversity of the data is an essential requirement of our study, we have selected three large, realistic, and freely available datasets described below.

The \textbf{RealWorld} dataset \cite{realword}, which contains 18 hours of recorded data, including accelerometer and gyroscope readings. Data were collected in 2016 from 15 subjects in 7 different device/body position configurations, using Samsung Galaxy S4 and LG G Watch R with a sampling rate of 50 Hz. The recording was performed outdoor, where the subjects were told to perform specific activities without any restraints. Eight activities have been labeled in the data: Climbing Down, Climbing Up, Lying, Sitting, Standing, Walking, Jumping, and Running. We believe the RealWorld dataset represents HAR data well in the wild and exhibits realistic high-class imbalance (for instance, the `standing' activity represents 14\% of the data while the `jumping' one is limited to 2\%). Furthermore, the dataset is particularly relevant for the study since a single participants' data can model each client as one would expect in realistic pervasive computing scenarios.s

% The dataset came with 

% \francois{please add the full duration of the below datasets and complete the missing information} 
% generating six minutes of data per device.

The \textbf{Heterogeneity Human Activity Recognition (HHAR)} dataset \cite{10.1145/2809695.2809718} accounts for 4.5 hours of recorded activities. \hl{Each of the 9 participants carried 8 Android smartphones (2 LG-Nexus 4, 2 Samsung Galaxy S3, 2 Samsung Galaxy S3 Mini, and 2 Samsung Galaxy S+) within a compact pouch and  4 Android smartwatches (2 LG watches and 2 Samsung Galaxy Gears) with 2 on each arm, were asked to perform five minutes of 6 activities (Biking, Sitting, Standing, Walking, Stair Up and Stair down). All 12 devices recorded the activities using the on-system accelerometer and gyroscope measurements to their respective maximum sampling rate (Various sampling ranges between 50 and 200 Hz). It is also interesting to mention that the authors of the dataset highlighted that the devices have gone through previous daily-life usage and hints of real-life induced usage errors on the recording device.}

Due to the variety and number of recording instruments, the HHAR dataset introduces a new evaluation perspective for FL algorithms on not just statistical heterogeneity but also system heterogeneity. \hl{To emphasize system heterogeneity, we combined the sensor data for each brand and treated them as one client (e.g., the two LG-Nexus 4 on subject one are considered as the data of a single client). % The two sets will be considered as two different clients to emphasize system heterogeneity. 
As a result, from the 6 unique devices and 9 clients, we generated 51 clients in total. Due to missing gyroscope data, 3 clients were discarded. Creating client from the same users reduces but does not erase the challenge of statistical heterogeneity since the devices where positioned at the same place on the body} \cite{realword}.

% generated a client per device for the six devices worn by each of the nine users. Due to missing gyroscope data, 3 clients were discarded, and 51 clients in total were used in our experiments. 
%Supposedly, the number of clients would be 54, but due to missing gyroscope data from three 3 clients \francois{(devices?)}, they were discarded and only 51 client were used in our experiments. 

The \textbf{Sussex-Huawei Locomotion (SHL)} Preview dataset {\cite{8418369}} is composed of 59 hours of annotated recordings acquired in 2017. It was recorded by 3 different users throughout their daily lives, each over a period of 3 days, with 4 different device/body positions. We only took the accelerometer and gyroscope data for our experiment while leaving out GPS, barometer, and magnetometer data for coherency purposes with the other dataset. The data were collected using only Huawei Mate 9 smartphones with a sampling rate of 100Hz. Each user was recorded in 3 distinct time sessions, with each session focused on different locomotions/activities. The SHL dataset has 6 labels: Still, Walking, Run, Bike, Car, Bus, Train, and Subway. Using the three different periods for each user data collection instance that came pre-segregated in the dataset, we generated 3 clients for each of the 3 users. The result is 9 clients for our simulations. \hl{However, the class distribution over the periods (i.e., clients) was severely imbalanced (e.g., subway and train were abundant in one client, but there were nearly no car labels). This imbalance is why a stratified split over the class labels was applied for each user data to compose clients with a balanced distribution (the reader is referred to} \cite{Usmanova2021} \hl{for promising solutions to deal with the class imbalance in an FL context).}  
%Although, as each client that were generated using the time instance had highly imbalanced class data (e.g., train and subway were abundant in one time span but there were nearly no car labels).
%We had thus did a balance stratified split over the class labels between the 3 clients for each user \hl{(user only or client ? -> evenly distributed within client ? )}to mitigate external problems such as forgetful learning.
%\hl{Hello Professor Francois! the class labels were evenly distributed (stratified partitioned the classes} for the 3 clients of every that were generated from each's user seperate timespan.

As some data were recorded with devices with different sampling rates, we had applied down-sampling on the collected data to a lower frequency of 50Hz. Such sampling frequency and data selection are in line with a survey on HAR on smartphones \cite{overviewHar} that has shown that the optimal sampling frequency is between 20 Hz and 50Hz and that accelerometers and gyroscopes are the most adequate sensors for classification.

\subsection{Evaluation strategy}\label{sec:eval_strategy}

%For the sake of reproducibility, all our experiments have been performed in simulation mode (like most federated learning evaluations). However, let us note that we have also implemented our approach on real devices, i.e., Google's Pixel 2, to check the approach's feasibility. This is not reported here due to a lack of space.
% this section is misleading abit, our baseline experiments were on mobile but our experiements with FL were simulations

As previously explained, the interest of FL over classical learning is the ability to merge several client models into a global one in order to improve genericity without degrading specialization. To assess these properties, we then decided to compute three different metrics for each experiment:
\begin{enumerate}
    \item \textbf{Global accuracy}. This accuracy is computed by the server with the global model aggregated from all the clients. It tells how well federated learning is able to create a general model and permits to answer the research question "\emph{Does FL bring better global performances than centralized global learning?}"
    \item \textbf{Personalization accuracy}. This accuracy is computed by the client using its local datasets. It tells how personalized the client models are and permits to answer the research question "\emph{Does FL bring better local performances with already seen data than local learning only?}"
    \item \textbf{Generalization accuracy}. This accuracy is computed by the client using the global dataset. It tells how well the client models are able to retain generalization and permits to answer the research question "\emph{Does FL bring better local performances with unfamiliar data than local learning only?}" This evaluation is crucial because it qualifies as one of the main potential benefits of the FL approach. 

\end{enumerate}

\begin{figure}[!htb]
\centering
\includegraphics[width=1.0\linewidth]{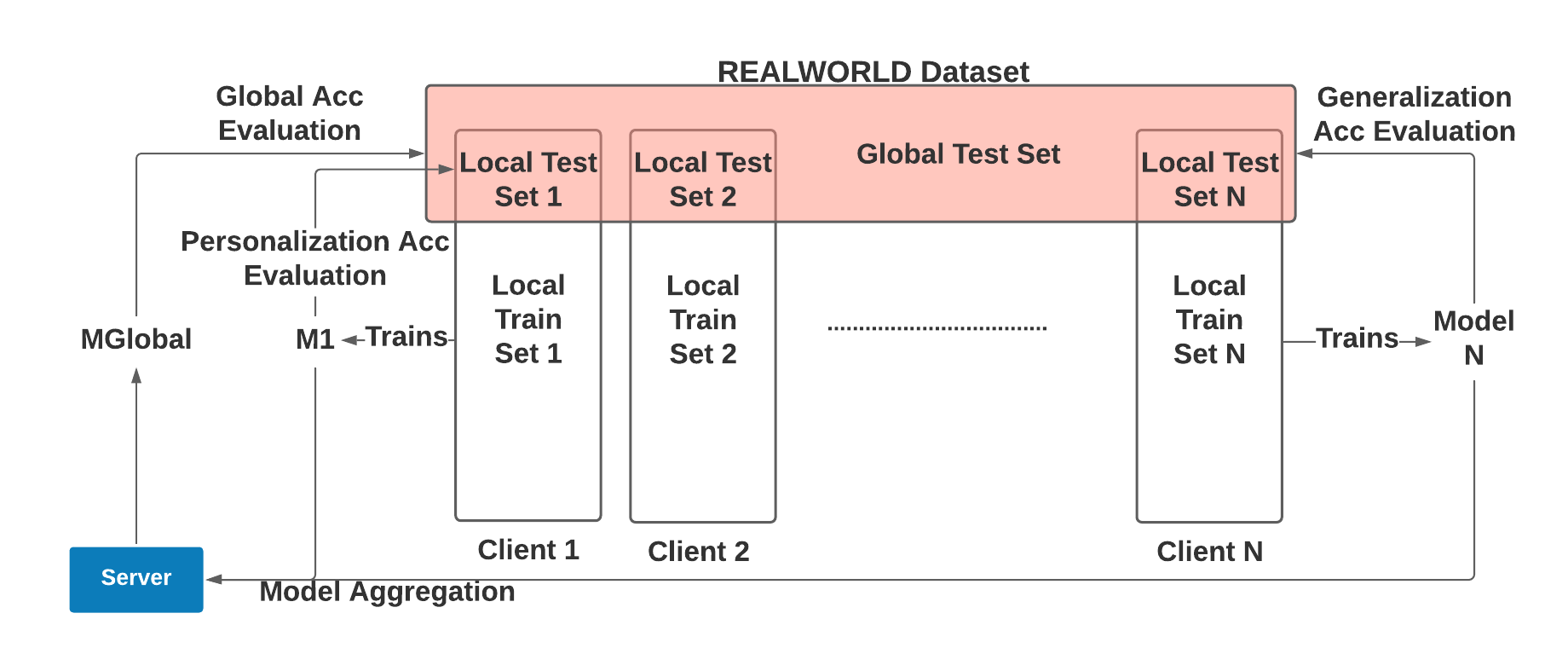}
\caption{Depiction of data partitioning/arrangement}
\label{fig:datasetBreak}
\end{figure}

\hl{Another important question of the evaluation is how to model clients. Many FL studies partition a unique dataset into several equally distributed datasets to represent the clients. This is not a realistic way of simulating heterogeneous clients. In our study on the RealWorld and HHAR datasets, each client is represented by unique traits such as a specific individual's activity habits, devices and device position.
% by the record of a single identified human participant without combining data from several participants. 
In that way, the FL algorithm is left dealing with a very personalized local model, which corresponds to realistic settings}. Figure~\ref{fig:datasetBreak} illustrates how partitioning has been designed. 
% Each client is represented by a set of records of one participant extracted from RealWorld.

For each client, the dataset is partitioned into a training set and a test set. The test set is used for the local evaluation of each client. These evaluation results are then aggregated to give the \textbf{Personalization accuracy} evaluation. The concatenation of all client test sets forms the global test set, which is used to evaluate the \textbf{Global accuracy}. It is also used for the local evaluation of each client. These evaluation results are then aggregated to give the \textbf{Generalization accuracy} evaluation. 
 
% Finally, a major question for applicability in the pervasive field is the ability to deal with dynamic clients that may leave or join the learning process at anytime. The study of federated learning with \emph{asynchronous} clients, to the best of our knowledge, has not been reported in the literature. In our evaluation, we consider an active/idle client pool to modulate the communication round client training. We start the training with an initial number of clients and randomly remove some of them and add some other every $n$ rounds. This evaluation is planned to shed light on the research question "\emph{how robust is FL to asynchronous training and varying number of clients?}"

In the rest of this work, for HAR performance metrics, we have used standard classification metrics such as accuracy, precision, and recall. From these, we have computed the F1 score as a complementary measure that gives a per-class evaluation, which is the harmonic mean of precision and recall. 

{In this study, in contrast with  conventional federated learning, where the goal is very server-centric and the objective is to optimize the server model for better Global accuracy, we seek a clients-centric optimization.  Our goal is that clients get well personalised while still able to deal with new situations. Hence, we mainly focus on improving the Personalization and Generalization accuracy.}

{Finally, it is worth noting that, beyond accuracy assessment, the level of heterogeneity of the learning environment can be observed through the gap between the Personalization and Generalization accuracy of the `Local' training setting. A large difference is the consequence of very heterogeneous clients (e.g. as with the HHAR dataset) while a small difference means that clients are homogeneous with one another (e.g. with the clients of the SHL dataset). }

\section{Experiment and Results} \label{sec:expe}

\subsection{Settings}

As said in Section~\ref{sec:datasets} the three datasets RealWorld, HHAR and SHL were respectively partitioned into 15, 51, and 9 clients. Each client dataset was, in turn, partitioned into an 80\% -- 20\% ratio to obtain local train and test datasets. Furthermore, the local test sets were aggregated into a global test set, used to evaluate the genericity/personalization trade-off of the different algorithms. 
The input data of the datasets were provided by the Inertial Measurement Unit (IMU) of smartphones. That is the 3-axis accelerometer data with the 3-axis gyroscope data (down)sampled at 50hz. As common in deep learning approaches (i.e., features are learned and not hand-crafted), no features extraction was applied to let the model build its own representation. The data was preprocessed using channel-wise z-normalization and sampled using a window-frame size of 128 with a 50 overlap of 6 channels of each axis.

Extending our evaluation against other FL algorithms (FedAvg, FedProx, and FedMA), we set up two additional experiment settings using conventional learning means. The \textbf{Centralized} experiment is where the same CNN model architecture is trained in conventional means where data from different clients are pooled together at the server. This experiment can be used as a comparison to evaluate how well FL algorithms produce a single generalized model. In addition, we have the \textbf{Local} learning settings where clients perform local training without any collaborative techniques. This setting can demonstrate the potential gain through collaborations when clients partake in the FL learning pool.

For each experiment, we performed three evaluations. In the \textbf{Global accuracy} evaluation, we tested the aggregated model against the combined global dataset. In the \textbf{Personalization accuracy} evaluation, we tested each client model on its own local dataset. In the \textbf{Generalization accuracy} evaluation, we take local models with the best personalization accuracy throughout the entire communication round and then test them on the combined global dataset. For each evaluation, we computed the accuracy, recall, precision, and F-score. 

All experiments were implemented using TensorFlow 2 with the FedAvg, FedMA, and FedProx implementations able to reproduce the results of the original authors. 

% and run on a Debian 4.19.132-1 version 10 using a GPU GeForce GTX TITAN Black 6GB. We used our own implementation of FedAvg and FedPer to overcome the limitations of TensorFlow Federated (TFF) (memory size) at the time of the experiment. FedMA was adapted from the own code of the authors with minimal modifications. FedDist was entirely implemented by us. The FedAvg, FedMA, and FedPer implementations were able to reproduce the results of the original authors. 

%Each client has been simulated using data corresponding to the dataset's individual participant, leading to 15 different clients. 
%The dataset activities are equally distributed to create a balanced client test scenario where all users have similar amounts of data.

%The RealWorld dataset is imbalanced with very little activity count for the ``jump''  activity compared to other activities. Consequently, the measurement with accuracy is computed as the number of correct classifications divided by the total number of instances. When the model may under-perform in classifying the ``jump'' activity, the information may not be delivered when the model’s performance is reported in accuracy.  Thus, we have employed the F1 score as a complementary measure that gives a per-class evaluation, which is the harmonic mean of precision and recall as our primary evaluation metric.

%\begin{figure}[!htb]
%\centering
%\includegraphics[width=\linewidth]{fig/localLearnLearningAccuracy.png}
%\caption{traditional learning over 200 epochs}
%\label{fig:localLearn}
%\end{figure}

\subsection{Initial HAR model with Traditional Learning}

We held several experiments to select a state-of-the-art HAR model where we trained models without any collaborative techniques. Following the study of \cite{ek:hal-03207411}, we selected, among other architectures, a 2-layered CNN model to limit the model complexity and size. The shape of this model is 196 filters of 16x1 convolution layer, followed by a 4x1 max pool layer, then 1024 units of the dense layer, and finally the softmax layer with 8 units. {The models for all experimented are trained using class weighted learning with an SGD optimizer.} This model reaches 92.48\% of F-Score, which is far above the current state-of-the-art on RealWorld dataset, which was previously 81\% of F-Score \cite{realword}. This model will be used in all the subsequent experiments. 

% To handle class imbalance from each client, as mentioned in the document, we use weighted training of classes for all the training instances.
% The models were trained during 200 epochs using a mini-batch SGD of size 32, and a dropout rate of 0.50 was employed.

% After the consideration of multiple studies and model hyper-parameters search, our choose to employ the 2 layered CNN

\subsection{Federated Learning Results}

The FL algorithms were run using the CNN model architecture defined in the previous section. Before learning, all models were randomly initialized. The four different algorithms were evaluated using the RealWorld dataset, while only FedAvg and FedDist were evaluated using HHAR and SHL datasets. 

\subsubsection{Results on the RealWorld dataset}

For RealWorld, the FL algorithms were trained for 200 communication rounds. The clients were all set to perform local learning for 5 epochs.  Table~\ref{fig:resultbl1} presents the results of the learning. The Centralized Server Global accuracy (92.48\%) and Local Personalization accuracy results (96.04\%) give the upper limit in terms of generalization and personalization. Among the FL algorithms, FedDist outperforms other algorithms for the three measures (Global, Personalization, and Generalization accuracy) with only the exception of the Global accuracy against FedProx. FedDist exhibits the best trade-off between generalization and personalization (Generalization accuracy = 74.23\%). FedAvg follows with a Generalization accuracy of 72.99\% with FedProx's 72.70\%. FedMA (60.09\%) shows a trend towards overfitting but without beating FedDist or FedAvg in personalization. In terms of convergence, FedProx quickly reached a peak at 110 rounds of communications and stopped obtaining gains in personalization accuracy afterward, while FedMA's Global accuracy gains stopped at the 137th communication round.
% Apart from FedMA, all algorithms seem slow in converging before the 200 communication rounds. We detail the results for each algorithm below. 

\begin{table}[!b]

\centering
\caption{Overall results of the learning experiments on the RealWorld dataset}

\resizebox{\columnwidth}{!}{
\begin{tabular}{|c|c|c|c|c|c|}
\hline

& \multicolumn{5}{c|}{ RealWorld } \\
\hline

& Global&Personalization& Generalization& Global  & Personalization \\ 
& F-Score (\%) & F-Score (\%) & F-Score (\%)  & Best Rnd  & Best Rnd\\ 

\hline

Centralized & \textbf{92.48} & N/A & N/A  & 197 (Epoch) & N/A  \\
Local & N/A & \textbf{96.04 $\pm$ 1.77 }& 51.94 $\pm$ 3.39 & N/A & 198 (Epoch) \\
FedAvg & 83.44 & 95.82 $\pm$ 1.53 & 72.99 $\pm$ 1.84 & 198 & 187  \\
{FedProx} & {85.87} & {94.55 $\pm$ 1.71} & {72.70 $\pm$ 2.36} & {193} & {110}  \\
% FedPer & N/A & 95.46 $\pm$ 1.62 & 53.01 $\pm$ 2.93 & N/A & 190 \\
FedMA & 78.67 & 93.65 $\pm$ 2.21 & 60.09 $\pm$ 1.73  & \textbf{137} & \textbf{171}  \\
FedDist & 84.52 & 95.84 $\pm$ 1.59 & \textbf{74.23 $\pm$ 2.29} & 196 & 191  \\
FedAvg$_{FedDist\_size}$  & 83.97 & 95.74  $\pm$ 2.35 & 73.73 $\pm$ 1.54 & 197 & 183  \\
\hline
\multicolumn{6}{r}{\small $\pm$ = Standard Deviation} \\
\end{tabular}
}

\label{fig:resultbl1}
\end{table}

FedAvg, despite its simplicity, seems difficult to beat. %As shown in Figure \ref{fig:fedAvgAcc}, with this method, 
At the server level it generalizes well on the global test-set with an F-Score of 83.44\% (which is still far from the centralized learning approach of 92.48\%). The client model obtained a mean F-Score of 95.82\% on the local-test set and 72.99\% on the global test set. 
%The learning curve showed in Figure~\ref{fig:fedAvgAcc} exhibits a classic shape where the learning starts with a steep improvement until it reaches a slow monotonic increase after communication round 20. It seems that the performance on the global-test set would be able to grow with more communication rounds further. We highlight that the slower convergence, compared to the traditional learning approach,  is due to the averaging property of FedAvg, where it produces effects similar to regularization techniques where we limit over-fitting at the cost of slower convergence. The client level learning is less monotonic with sometimes sharp changes in the standard deviation for the Personalization and Generalization accuracy evaluation. This trait can be attributed due to some clients' peculiarities. 

FedProx presents results very representative to its nature, where training is done with respect to the server. The algorithm has the best Global accuracy (85.87\%) compared againts other FL algorithms. The drawback, although, is that FedProx losses in Personalization accuracy (Personalization accuracy = 94.55\%) and Generalization accuracy againts FedAvg (Generalization accuracy = 72.70\%). 
% With the FedPer algorithm, only the CNN layer part is communicated to the server while the dense layer (i.e., the personalized layer) is kept local. This scheme might explain why the FedPer algorithm does not succeed in generalizing (Generalization accuracy = 53.01\%). %, as it can be seen in Figure \ref{fig:fedPerAcc}. 
% The personalization layer stays too strong, and such the model is not able to react appropriately to new data. However, the results show that even the personalization feature of FedPer is not significantly better than any of the other FL algorithms.
%We found that the clients using the CNN on the RealWorld dataset can retain their ability to personalize and perform well, with 95.46\% macro F-Score on their local test-sets. The exhibited client’s F-Score of 53.01\% on the global test-set suggests that FedPer provides little advantage regarding the client’s model ability to generalize.

%\begin{figure}[!htb]
%\centering
%\includegraphics[width=\linewidth]{fig/fedPerRWCNNLearningAccuracy.png}
%\caption{FedPer learning over 200 communication round}
%\label{fig:fedPerAcc}
%\end{figure}

FedMA reached an F-Score of 78.67\% for the Global accuracy. %, as exhibited in Figure \ref{fig:fedMAAC}. 
It has the lowest ability to personalize with some minor drawbacks (Personalization accuracy = 93.65\%) while performing moderately on the global test-set (Generalization accuracy = 60.09\%). Furthermore, FedMA has a much higher training cost. Indeed, in our experimentation, 5 local epochs are used to train client models. For FedMA, this means a total of 25 local epochs for each communication round (5 local epochs for each of the first and second layer, and another 15 for the softmax layers).

%\begin{figure}[!htb]
%\centering
%%\includegraphics[width=\linewidth]{fig/fedMACNNRWLearningAccuracy.png}
%\caption{FedMA learning over 200 communication round}
%\label{fig:fedMAACC}
%\end{figure}

FedDist presents the best performance overall in our experiments. The learning curve shown in Figure~\ref{fig:fedDistAC} exhibits a two-step behavior. This is because at a certain point, after many communication rounds, no new neuron or filter is added to the global model, and thus, training can stabilize better onward. This behavior is further enforced with a penalty function that takes the current communication round as input to raise new neurons' acceptance rate in later rounds. After no new neurons are needed, we expect the late stages of training similar to the FedAvg algorithm. However, the initial step was sufficient to present better performances than FedAvg. In the end, FedDist gets the best Generalization accuracy (e.g., the best trade-off between generalization and personalization) and could even reach more with further training.  
%
%\hl{In terms of training time and communication cost, depending on the number of layers in the chosen model, FedDist initially requires certain folds more communication than FedAvg. This is because of the layer-wise training, as used in FedMA. Although contradicting FedMA's approach, all the layers (including the softmax) use the same number of local epoch and will skip layer-wise training for a specific layer if no neuron was added in the preceding layer. Due to this scheme and the lesser aggregation complexity of FedDist, our algorithm requires less training time and communication against FedMA.}
%
{The final CNN model for FedDist has grown from 196 to 222 filters for the 1st layer (growth of 1.13) and from 1024 neurons to  2246 (growth of 2.19). This is comparable to the experiments of FedMA\footnote{\hl{For a fair comparison with the other algorithms and to reduce cost, FedMA was trained with the default hyper-parameters reported in} \cite{wang2020federated}. \hl{An adapted tuning might lead to different results}.} for which the growth rate can go from 1.004 in the top layers to 2.64 in the bottom layers.}

%In the final global model shape of the 2 layered CNN, from the original 196 filters in the first convolutional layer, 26 new filters have been added. In the original 1024 neurons in the second layer, 1222 new neurons were gained.

{In order to investigate the performance gains of FedDist beyond the growth of the model shape, we have taken the final model shape that FedDist obtained and reinitialized it to start again with FedAvg (FedAvg$_{FedDist\_size}$ in Table~{\ref{fig:resultbl1}}). Table}~ \ref{fig:resultbl1} {shows that although FedAvg$_{FedDist\_size}$ has better results than FedAvg, it is still short of the original FedDist implementation. We pinpoint two possible explanations. First, the neurons that FedDist adds are specialized instead of randomly initialized trained neurons. Second, due to the layer-wise training FedDist, neurons in the upper layers of the model receive more training as they adjust to new neuron weights that come from lower layers. }

\subsubsection{Results on the HHAR and SHL datasets}

\begin{table}[!b]

\centering
\caption{FL F-score and standard deviation for the HHAR and SHL datasets}

\resizebox{\columnwidth}{!}{
\begin{tabular}{|c|c|c|c|c|c|}
% \hline
% & Global&Personalization& Generalization& Server  & Client \\ 
% & F-Score (\%) & F-Score (\%) & F-Score (\%)  & Best Rnd  & Best Rnd\\ 
% \hline 
% \hline
% & \multicolumn{5}{c|}{ HHAR Dataset } \\
% \hline
% Centralized & \textbf{90.37} & N/A & N/A  & 199 (Epoch) & N/A  \\
% Local & N/A & \textbf{93.21 $\pm$ 11.86 }& 25.59 $\pm$ 9.24 & N/A & \textbf{ 146 (Epoch)} \\
% FedAvg & 73.27 & 90.14 $\pm$ 11.13 & 57.86 $\pm$ 11.37 & 199 & 196  \\
% FedDist & 75.02 & 91.35 $\pm$ 12.72 & \textbf{62.06 $\pm$ 11.43} & 198 & 196  \\
% \hline
\hline
& Global&Personalization& Generalization& Global  & Client \\ 
& F-Score (\%) & F-Score (\%) & F-Score (\%)  & Best Rnd  & Best Rnd\\ 
\hline 
\hline
& \multicolumn{5}{c|}{ HHAR Dataset } \\
\hline
Centralized & \textbf{96.74} & N/A & N/A  & 197 (Epoch) & N/A  \\
Local & N/A & 94.64 $\pm$ 13.57 & 31.47 $\pm$ 9.65 & N/A & 198 (Epoch) \\
FedAvg & 80.7 & 93.91 $\pm$ 11.3 & 69.47 $\pm$ 6.41 & 200 & 197  \\
FedDist & 82.81 & \textbf{95.52 $\pm$ 9.51} & \textbf{73.66 $\pm$ 7.71} & 198 & 196  \\
{FedAvg$_{FedDist\_size}$} &80.36  & 93.58 $\pm$ 11.45 & 67.78 $\pm$ 6.74  & 200 & 167  \\
\hline
\hline
& \multicolumn{5}{c|}{ {SHL Dataset} } \\
\hline

Centralized & \textbf{88.89} & N/A & N/A  & 197 (Epoch) & N/A  \\
Local & N/A & \textbf{88.85 $\pm$ 1.68 }& 50.39 $\pm$ 1.26 & N/A & 199 (Epoch) \\
FedAvg & 76.37 & 86.79 $\pm$ 2.30 & \textbf{66.62 $\pm$ 1.37} & 198 & 187  \\
FedDist & 76.37 & 86.79 $\pm$ 2.30 & \textbf{66.62 $\pm$ 1.37} & 198 & 187  \\
{FedAvg$_{FedDist\_size}$}& 76.37 & 86.79 $\pm$ 2.30 & \textbf{66.62 $\pm$ 1.37} & 198 & 187  \\
\hline

\end{tabular}
}
\label{fig:resultblHHAR}
\end{table}

{In the following, only FedAvg and FedDist of the FL implementations algorithms were evaluated on the HHAR and SHL datasets since FedMA and FedProx were not found to be competitive in Personalization and Generalization. FedMA in our heterogeneous learning environment could not fix a new architecture shape that matched the collaborative learning need. FedProx, in its nature, is very server model-centric, while the theme of our works is a client-centric one where we emphasize the user model's Personalization and Generalization gains.}
On the highly diverse and client intensive (51) HHAR dataset, the results shown in Table~\ref{fig:resultblHHAR} exhibit that local training reached the competitive personalization score (94.64\%), but the generalization score fell far behind the FL methods by a considerable margin (Local 31.47\% vs. FedAvg 69.47\% vs. FedDist 73.66\%). This is clearly due to the client's heterogeneity which also explains the large standard deviation values. Regarding the global score, the centralized approach here has a strong F-score of 96.74\% (against FedAvg's 80.70\% and FedDist 82.81\%). Here, FedDist was still able to outperform FedAvg with a 1.6\% increase in personalization score and a near 4\% increase in generalization score. Surprisingly, the local personalization score which has the best in all other experiments, was beaten by FedDist. The two-layered CNN had grown substantially, with 257 filters in the 1st convolutional layer and 8558 neurons in the 2nd dense layer. {Despite this growth FedAvg$_{FedDist\_size}$ was not able to benefit from the higher capacity and exhibits performances even below the original FedAvg.} The study on the HHAR dataset exhibits the need for better system heterogeneity management in FL methods. Federated learning should not only take into account the idiosyncratic-ness of user behavior but should also address the difference between users' hardware.
For the SHL Dataset, FedAvg and FedDist give the same results for all three metrics (hence FedAvg$_{FedDist\_size}$ also gives the same results). This is due to the class-stratified distribution done to generate uniform clients. Here FedDist behaves as expected; when client models are very similar to one another, no new neuron will be added to the model. This shows that when client models are sufficiently similar, FedDist behaves as FedAvg. 

%\sannara{On the client plenty and highly diverse HHAR dataset, a good gain in Generalization accuracy for the federated learning methods is observed. Despite the Local training approach having the best personalization accuracy (93.21\%), the Generalization accuracy fell behind the federated learning methods by a considerable margin (Local 25.59\% vs. FedAvg 57.86\% vs. FedDist 62.06\%). Furthermore, we can note the large standard deviation values for the distributed learning methods in this case study. This cause can be inferred from the numerous client count that varies in performance, where few clients are under-performing on personalization. On the Global accuracy, the centralized approach here has a strong F-score of 90.37\% against FedAvg's 73.27\% and FedDist 75.02\%. Additionally, this experiment showed that the Centralized approach Global accuracy was even able to outperform the Personalization accuracy of FedAvg (90.37\% vs. 90.14\%). This here can reduce the need for federated learning for when user data can be communicated. }

\subsection{Asynchronous Clients}

\begin{figure}[!bt]
    \centering
    \begin{subfigure}{.5\textwidth}
        \centering
        \includegraphics[width=0.95\linewidth]{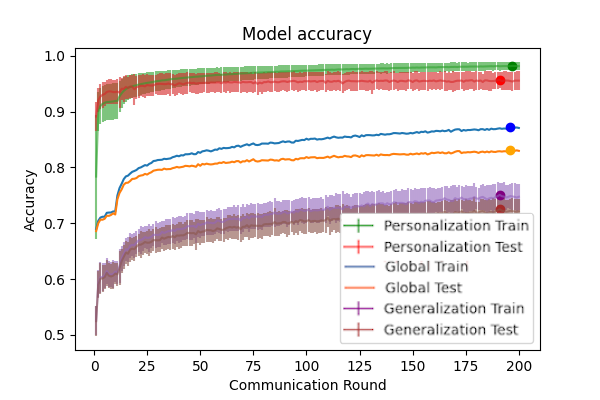}
        \captionof{figure}{FedDist learning over 200 CRs on the RealWorld dataset}
        \label{fig:fedDistAC}
    \end{subfigure}%
    \begin{subfigure}{.5\textwidth}
        \centering
        \includegraphics[width=0.84\linewidth]{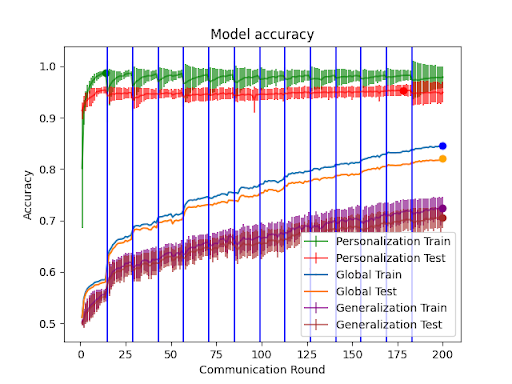}
        \captionof{figure}{FedDist in an incremental clients asynchronous environment}
        \label{fig:fedDistAsynIncreasing}
    \end{subfigure}
    \begin{subfigure}{.5\textwidth}
        \centering
        \includegraphics[width=0.9\linewidth]{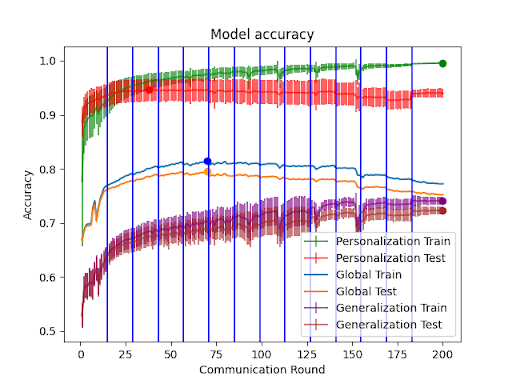}
        \captionof{figure}{FedDist in an decremental clients asynchronous environment}
        \label{fig:fedDistAsynDecreasing}
    \end{subfigure}%
        \begin{subfigure}{.5\textwidth}
        \centering
        \includegraphics[width=0.9\linewidth]{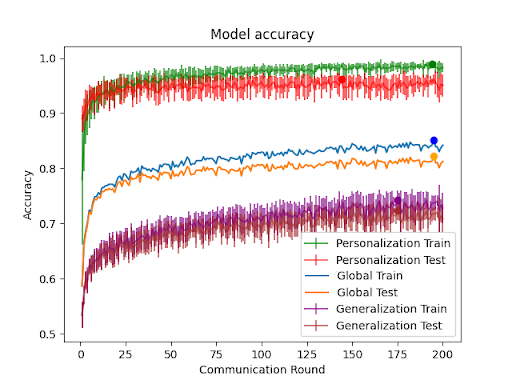}
        \captionof{figure}{FedDist in an interchanging clients asynchronous environment}
        \label{fig:fedDistAsyncHalf}
    \end{subfigure}%
    \caption{The asynchronous simulations in 3 different environments}
\label{fig:fedDistAsync}
\end{figure}

% \begin{figure}[!htb]
% \minipage{0.32\textwidth}
%   \includegraphics[width=1.0\linewidth]{fig/increasing.png}
%   \caption{A really Awesome Image}\label{fig:awesome_image1}
% \endminipage\hfill
% \minipage{0.32\textwidth}
%   \includegraphics[width=\linewidth]{fig/decreasing.png}
%   \caption{A really Awesome Image}\label{fig:awesome_image2}
% \endminipage\hfill
% \minipage{0.32\textwidth}%
%   \includegraphics[width=\linewidth]{fig/half.png}
%   \caption{A really Awesome Image}\label{fig:awesome_image3}
% \endminipage
% \end{figure}

%If performance during simulation is essential to understand FL algorithms behavior, this does not inform the ability of such algorithms to deal with real-world constraints. For instance, 
In pervasive computing, clients are mobile and can be disconnected/reconnected anytime. Hence, FL must be robust to such situations where the consequences can be detrimental to the learning. To study this behavior, we implemented an active/idle client pool to modulate the number of clients at each communication round. {Such study also provides insight regarding the behavior of the method with respect to unseen users.} Using the 15 subjects of the RealWorld dataset, we experimented with three different asynchronous environment scenarios: \textbf{Incrementing clients} (the training starts with 2 clients, and every 14 rounds, a new client is added to the active client pool), \textbf{Decrementing clients} (the training starts with 15 clients and every 14 rounds, a client is removed), \textbf{Interchanging clients} (at every round, 8 clients are randomly selected).

The effect of the three simulations is shown in Figure~\ref{fig:fedDistAsync}. In the incrementing case (Figure~\ref{fig:fedDistAsynIncreasing}), the Global and Generalization score improves as more clients come in at each of the 14 rounds intervals (blue lines on the figures). However, when a new client joins the active pool, the active clients exhibit a sharp variation in their Personalization score (green curves). This is expected since active clients must adapt to the newly added model. However, the clients could all recover to well-performing scores within a few communication training rounds.
%that changes the server model's convergence direction during the aggregation stage. The new server model causes slight but not drastic adverse effects, and the clients were all able to recover to well-performing accuracy scores within a few communication rounds of training.
In the decrementing case (Figure~\ref{fig:fedDistAsynDecreasing}), the Personalization and Global score are constantly decreasing as more and more clients leave. %This finding hints that clients are able to help one another do better on their local data when in a collaborative learning framework. 
An interesting behavior is that Generalization score is overall gaining in performance despite the removal of clients from the active pool and obtains a generalization level reasonably close to non-asynchronous experiments (reaching ~70\%). This finding shows that the remaining client can still benefit from the previous client's model that has left earlier. This phenomena hints specialized neurons obtained being able to retain previously learned features for continual learning purposes.  %Although, the server model suffers the most in Global accuracy to where performance slowly approaches the same level as the generalization accuracy. 
In addition, the simulation suggests that the server model can retain knowledge from previous clients that are no longer in the active training pool.
Finally, in the Interchanging case (Figure~\ref{fig:fedDistAsyncHalf}), we can see training curves with the same trends as in the classical case (Figure~\ref{fig:fedDistAC}) but with more variation due to the constant fluctuation of clients in and out of the active training pool. %Despite the frequent oscillation, a distinct gain in performance for all three metrics is seen that points to a slow convergence to the same performance results as seen in the non-asynchronous experiments. 
The results here show that clients do not need to be present at every communication round to contribute to and benefit from FL.

% From the client side, on the contrary, the new global model causes slight but not drastic adverse effects to client own. However, as training goes one it is quickly fixed. This can be explained by looking at Figure~\ref{fig:asyncAccAll} which shows the average distance and its standard deviation between all clients and the global model (cf. section~\ref{sec:metrics}). Each time a new client/model arrives, the its distance with the global model increases but the algorithm is able to make this distance converges within the final range by the end of the 20-round interval. 

% \begin{comment}
% \begin{figure}[!tb]
% \centering
% \includegraphics[scale=0.43]{fig/asyncLayerClientEuclid.png}
% \caption{FedAvg euclidean distance by layers }
% \label{fig:asyncAccLayer}
% \end{figure}
% \end{comment}

% \section{Evaluation Method}\label{sec:eval}

\section{Conclusion and further work}\label{sec:conclu}
Federated Learning (FL) exhibits clear theoretical advantages over classical centralized learning from a pervasive computing perspective. It provides a solution for distributed learning and, to some extent, privacy preservation. However, little is known about the behavior of such a learning approach and how to evaluate it in realistic pervasive computing situations. 
%Up to now, most of the studies about FL have been conducted in the computer vision area, without considerations for the specific needs of pervasive applications. Furthermore, 
Despite the fact that FL is becoming an active research area, our study reveals that the standard simple FL algorithm FedAvg is difficult to beat and that recent complex algorithms do not demonstrate a clear superiority. %In this paper, our new FL algorithm FedDist unites the efficiency of FedAvg with the capability of adapting the model architecture during training to handle data heterogeneity. 
%We also introduce a straightforward methodology to evaluate state-of-the-art FL algorithms and FedDist in the context of a HAR from smartphone sensors.  
%
%
% \begin{figure}[!tb]
% \centering
% \includegraphics[scale=0.43]{fig/asyncallClientEuclid.png}
% \caption{FedAvg euclidean distance by clients}
% \label{fig:asyncAccAll}
% \end{figure}
%
However, our evaluation method showed that our new FL algorithm FedDist outperformed the other FL algorithms on the three measures of generalization and personalization. Indeed, FL should lead to a high degree of adaptation to the device while keeping a high degree of generalization (e.g., prevent over-fitting and high client accuracy on global data). FedAvg also exhibited such behavior, but more complex FL algorithms such as FedMA and FedPer could not keep a high degree of generalization and did not show a superior capacity for personalization. On another note, choosing the optimal number of neurons for each layer of a NN is an open problem. FedDist relaxes the need to perform intensive hyper-parameter searches for the best model shape through the algorithm's ability to compute specialized neurons that are appended to the global model architecture. Finally, our study shows that FedDist exhibited good robustness to asynchronous learning when devices come and go during the learning.

% One advantage of FedDist over other algorithms is its ability to make the initial CNN model evolves along with communication rounds. Since deciding on the initial size of a NN model is an open problem, the property FedDist computing the number of new specialized neurons to add automatically to a model provides a flexible way to adapt the model architecture to the task. 
%While FedDist is more computationally intensive than FedAvg, it is far less complex than the FedMA algorithm and with better performance on HAR.

%Although these results add credence to the interest of federated learning for pervasive computing, a lot of challenges still remain. The study must be replicated with more datasets and different tasks\cite{BRENON201892}. We also plan to study the robustness of FL in scenarios such as asynchronous learning (devices come and go), a sudden change in client data, communication issues, heterogeneous population of devices (e.g., traveling device), and mismatches between server data and clients (noisy acquisition). 
\hl{Although these results show the interest of FL for pervasive computing, many challenges remain. For example, long-term studies are needed to optimize communication schedules and deal with lifelong learning} \cite{chen2018lifelong}. \hl{For example, our study did not examine the effect of extreme variations in class distribution across time and clients. Some preliminary work on the topic using continuous learning suggests that such a case is difficult} \cite{Usmanova2021}. \hl{Furthermore, due to the lack of large datasets from different users, FedDist has yet to be evaluated on a large scale. This requires a community effort to build large datasets and establish benchmarks for comparison and replication of research in this area.}

%% The Appendices part is started with the command \appendix;
%% appendix sections are then done as normal sections
%% \appendix

%% \section{}
%% \label{}

%% If you have bibdatabase file and want bibtex to generate the
%% bibitems, please use
%%
\bibliographystyle{elsarticle-num} 
\bibliography{bibfile.bib}
% \bibliographystyle{elsarticle-num} 
%%  \bibliography{<your bibdatabase>}

%% else use the following coding to input the bibitems directly in the
%% TeX file.

\end{document}